\documentclass{article}
\usepackage{ijcai24}

\usepackage{times}
\usepackage{soul}
\usepackage{url}
\usepackage[hidelinks]{hyperref}
\usepackage[utf8]{inputenc}
\usepackage[small]{caption}
\usepackage{graphicx}
\usepackage{amsmath}
\usepackage{amsthm}
\usepackage{booktabs}
\usepackage{algorithm}
\usepackage{algorithmic}
\usepackage[switch]{lineno}
\usepackage{epsfig}
\usepackage{amssymb}
\usepackage{color, colortbl}
\usepackage{tipa}
\usepackage{times}
\usepackage{epsfig}
\usepackage{amssymb}

\def\ie{$i.e.$}
\def\eg{$e.g.$}

\title{Bridge to Non-Barrier Communication: Gloss-Prompted Fine-grained Cued Speech Gesture Generation with Diffusion Model}

\author{
Wentao Lei$^{1, 2}$
\and
Li Liu$^{1, 3}$\thanks{Corresponding Author: avrillliu@hkust-gz.edu.cn.}\and
Jun Wang$^2$
\affiliations
$^1$The Hong Kong University of Science and Technology (Guangzhou)\\
$^2$Tencent AI Lab\\
$^3$The Hong Kong University of Science and Technology\\
}

\begin{document}

\maketitle
\begin{abstract}
Cued Speech (CS) is an advanced
visual phonetic encoding system that integrates lip reading with hand codings, enabling people with hearing impairments to communicate efficiently. CS video generation aims to produce specific lip and gesture movements of CS from audio or text inputs. The main challenge is that given limited CS data, 
we strive to simultaneously generate fine-grained hand and finger movements, as well as lip movements, meanwhile the two kinds of movements need to be asynchronously aligned.
Existing CS generation methods are fragile and prone to poor performance due to template-based statistical models and careful hand-crafted pre-processing to fit the models.
Therefore, we propose a novel \textbf{Gloss}-prompted \textbf{Diff}usion-based CS Gesture generation framework (called \textbf{GlossDiff}). Specifically, to integrate additional linguistic rules knowledge into the model. we first introduce a bridging instruction called \textbf{Gloss}, which is an automatically generated descriptive text to establish a direct and more delicate semantic connection between spoken language and CS gestures. Moreover, we first suggest rhythm is an important paralinguistic feature for CS to improve the communication efficacy. Therefore, we propose a novel Audio-driven Rhythmic Module (ARM) to learn rhythm that matches audio speech. Moreover, in this work, we design, record, and publish the first Chinese CS dataset with four CS cuers.
Extensive experiments demonstrate that our method quantitatively and qualitatively outperforms current state-of-the-art (SOTA) methods. We will release code and data at \href{https://glossdiff.github.io/.}{https://glossdiff.github.io/.}
\end{abstract}


\section{Introduction}
\begin{figure*}[htb]
 \centering
 \centerline{\includegraphics[width=16.0cm]{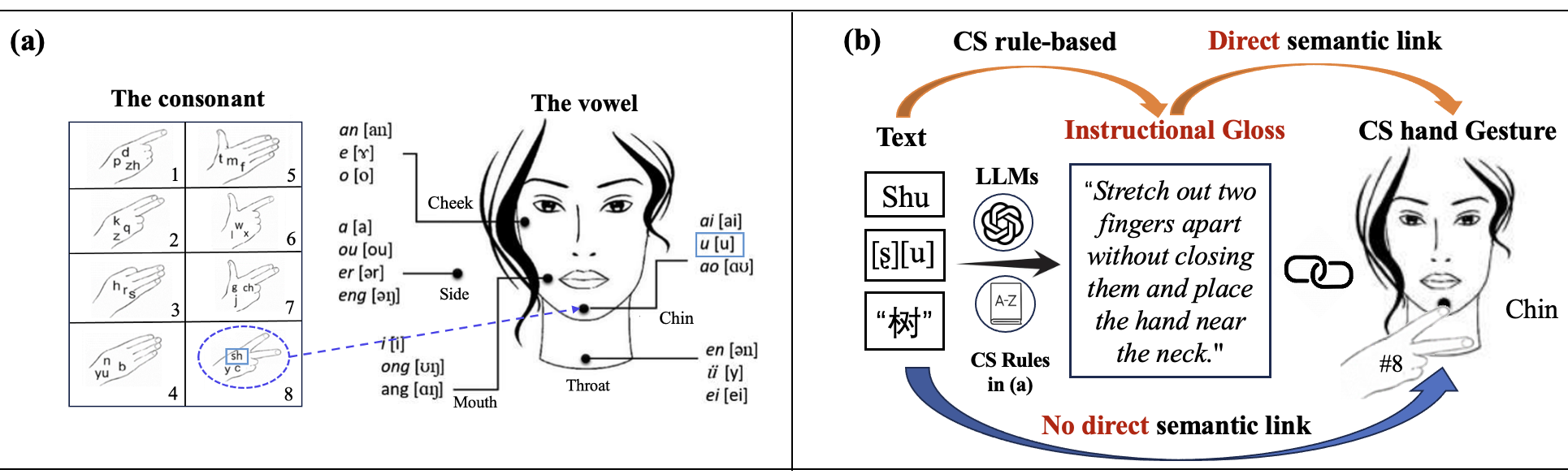}}
\caption{The details of CS rules and conversion process. (a) is the chart for the Mandarin Chinese Cued Speech (figure from [3]),
where five different hand positions are used to code vowels, and eight finger shapes are used to code consonants in Mandarin Chinese. (b) shows the proposed instructional gloss, which directly links the text to the CS movements.}
\label{fig:gloss}
\end{figure*}


According to the World Health Organization (WHO), more than 5\% of the global population (466 million) suffers from the hearing loss. 
As a predominant communication method for hearing-impaired people, Lip reading \cite{lip1,lip2} has a major defect of visual confusion.
For instance, it struggles to differentiate pronunciations with similar labial shapes, such as [u] and [y], posing challenges for hearing-impaired individuals in accessing spoken language through conventional education. 

To tackle the limitations of lip reading, and to improve the reading skills of individuals with hearing impairments, in 1967, Cornett introduced the \textbf{Cued Speech (CS)} system \cite{cornett1967cued}, which employs several hand codings (\ie, finger shapes and hand positions) to complement lip reading, providing a clear visual representation of all phonemes in spoken language \cite{lip1,lip2}. For instance, in Mandarin Chinese CS (MCCS) \cite{pilot} (see Fig. \ref{fig:gloss}(a)), it utilizes five hand positions for encoding vowel groups and eight finger shapes for encoding consonant groups. With CS, individuals with hearing impairments can differentiate sounds that might appear similar when observed on lips by incorporating hand information. Another widely adopted communication method is Sign Language (SL) \cite{sl1,sl2,sl3}. It is crucial to emphasize that CS is not a visual language like SL; instead, it is a coding system of spoken language \cite{cornett1967cued}. 
In addition, studies indicate that CS can be learned much more quickly than SL \cite{reynolds2007examination}.  
Given that CS can effectively promote non-barrier communication, audio/text to CS gestures video generation draws researchers' attention. It should be noted that comparing to text, CS is more friendly and more easily adopted by the hearing impaired who are illiterate \cite{cox2002tessa,power2007german}.

The multi-modal CS gesture generation is a challenging task for the following reasons: 1) high requirement for fine-grained and accurate gesture generation, as shown in Figure 1(a), where nuances in the hand's position and fingers' shape lead to quite different semantic meanings; 
2) the limited size of CS datasets and expensive annotation cost of complicated fine-grained CS gestures.
To address these challenges, we design a novel \textbf{Gloss}-Prompted \textbf{Diff}usion-based CS Gesture generation framework (\textbf{GlossDiff}). Specifically, 
we first propose a CS gloss, which is a direct motion instruction for bridging the gap between spoken language and CS gestures. It is automatically generated by LLM based on the encoding rule of CS in Figure \ref{fig:gloss}(a).
As shown in Figure \ref{fig:gloss}(b), when expressing the word ``tree", which is pronounced as ``/$\textit{\textipa{\textrtails}}$/ /u/" in Chinese, we generate the intermediate instruction text (\ie, gloss) to describe the process of using CS gestures to express this word, \ie, \textit{Stretch out two fingers apart without closing them and place the hand near the neck.}
Besides, we design a Gloss-Prompted Diffusion Model that can generate accurate hand and finger movements.

Moreover, rhythm is a critical paralinguistic information in spoken language. As a coding system for spoken languages, we suggest natural rhythm dynamics should also be considered as a very important feature for CS's complete semantic expression. More specifically, the rhythm here refers to the ability to generate multi-modal CS speech gesture movements (\ie, hand and finger movements), which match the phoneme durations and utterance prosody of speech. Unfortunately, previous works have not pay enough attention to this.
To this end,  we propose an \textbf{A}udio-driven \textbf{R}hythmic \textbf{M}odule (\textbf{ARM}) that considers the overall rhythm of the CS movements aligning with speech signals. We leverage the large-scale WavLM \cite{chen2022wavlm} to extract audio features, which we demonstrate outperforming traditional MFCC features.



We summarized our contributions as follows: \textbf{1)} A novel GlossDiff framework that simultaneously generates fine-grained hand position, finger movements, and lip reading in CS. Specifically, we introduce a CS \textbf{gloss}, which establishes a direct link between text/audio and CS hand movements, enabling more specific prompts for an accurate fine-grained CS gesture generation. \textbf{2)} A new module \textbf{ARM} that improves the overall rhythm of the CS movements. \textbf{3)} Publication of the first multi-cuer large-scale Mandarin Chinese CS (\textbf{MCCS}) dataset, which contains four cuers\footnote{The people who perform CS are called the cuer} and 4000 CS videos. \textbf{4)} Extensive experiments conducted on the MCCS dataset show the proposed GlossDiff achieves SOTA performance under different metrics. The qualitative and ablation studies, as well as user studies, further verify the effectiveness of the proposed model.

\begin{figure*}[htb]
\begin{minipage}[b]{1.0\linewidth}
 \centering
 \centerline{\includegraphics[width=15.0cm]{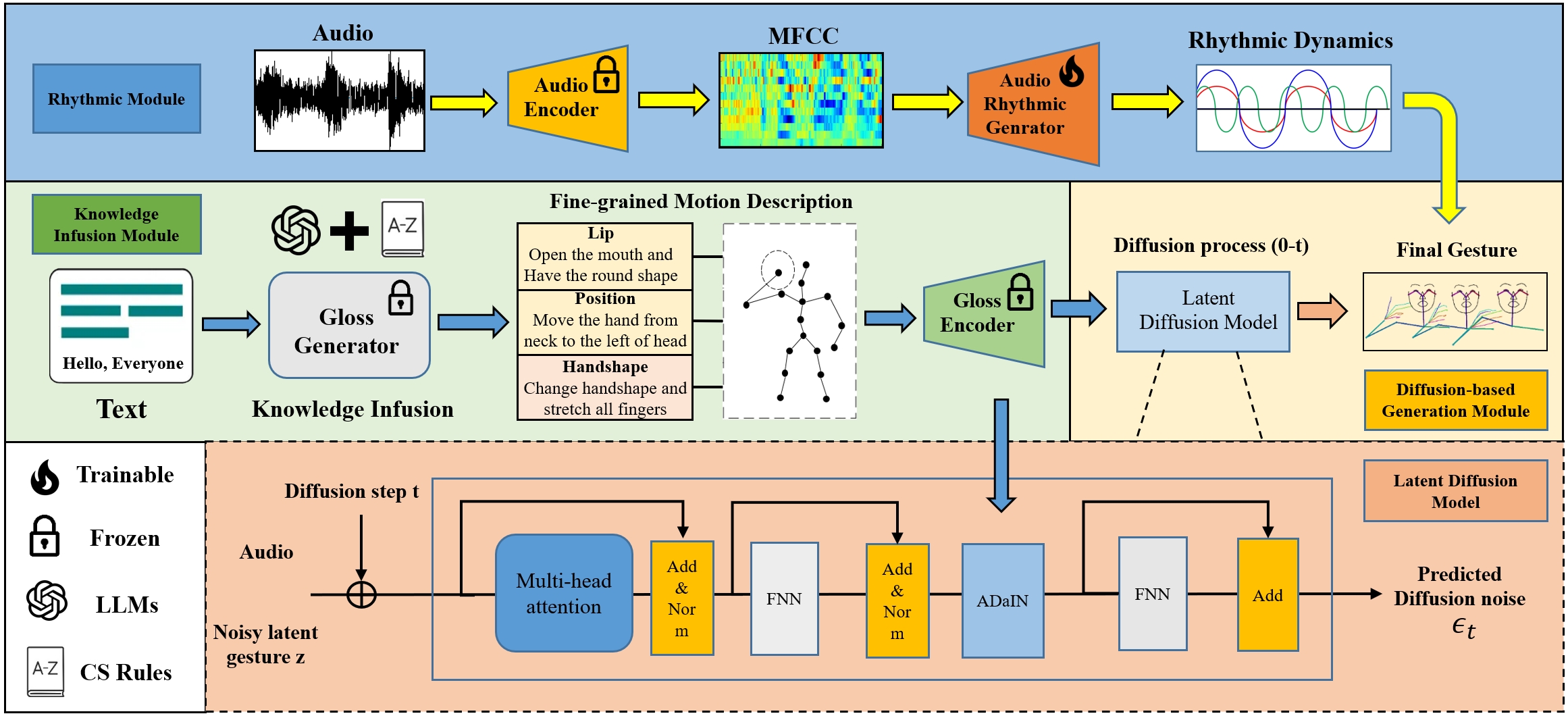}}
\end{minipage}
\caption{The overall framework of the proposed GlossDiff, where (a), (b), (c) represent the Knowledge Infusion Module, Audio Rhythmic Module and Diffusion-based generation module, respectively.}
\label{fig:cuers}
\end{figure*}

\section{Related work}


\subsection{Cued Speech Generation}

In prior work, early attempts at CS gesture generation \cite{csg1,csg2} are mainly rule-based. Notably, in \cite{csg1}, specific keywords were manually selected, along with low-context sentences \cite{IEEE}, and manual templates for corresponding hand gestures were predefined. This processing involved CS recognition, followed by mapping the recognized text to the hand templates. However, this method relied heavily on hand-crafted designs, which constrained the expressiveness of CS gestures and increased the required manual effort. In \cite{csg2}, a post-processing algorithm was introduced to refine synthesized hand gestures, including adjustments for hand rotation and translation. However, this approach required prior human knowledge to adapt the algorithm to new images, resulting in limited robustness.
To the best of our knowledge, there is still a gap in research about end-to-end deep learning-based CS gesture generation.

\subsection{Co-speech and Sign Language Generation}

The generation of Co-speech gestures involves generating body movements corresponding to audio input.
Previous studies mainly developed large speech-gesture datasets to learn how speech audio maps to human skeletons using deep learning, as in \cite{RG}. 
To make gestures more expressive, some methods use Generative Adversarial Networks (GANs) for more realistic results \cite{ginosar2019learning,GTC}. Recently, diffusion models like DiffGesture \cite{zhu2023taming}, effectively links audio and gestures while keeping time consistency, allowing for high-quality Co-speech gestures.
However, Co-speech gesture generation focuses on fluency and style rather than gesture fine-grained accuracy. Existing methods cannot generate accurate subtle CS hand gestures.

In the literature, there are several Sign Language (SL) generation methods: 1) The Neural Machine Translation approach from \cite{text2gesture} sees SL generation as translation, using neural models to process SL text. 2) The Motion Graph method in \cite{text2gesture} uses motion graphics to make a directed graph from motion capture data for SL creation. 3) Conditional generation methods, like GANs and VAEs, are also used for SL gestures. 4) Some researches have introduced transformer-based models for SL, as mentioned in \cite{DBLP:journals/corr/abs-2004-14874}.
Despite these advancements in SL gesture generation, applying these methods to CS gestures has limitations. Firstly, CS gesture generation necessitates more precise methods to achieve complex fine-grained gesture generation, while SL gesture generation is more coarse-grained. Secondly, SL gesture is not related with lip-reading, thus cannot match the speech rhythm and gesture-speech asynchrony characteristics \cite{re3,liu2023cross} in CS gesture generation.  

\subsection{NeRF and Diffusion-based Gesture Generation}
NeRF is a novel technique in 3D modeling, which effectively creates highly detailed and photo-realistic static scenes from 2D images. Its application has been extended to generating life-like talking head models \cite{guo2021ad}, demonstrating NeRF's capability in handling subtle facial movements and expressions. However, the application of NeRF in full-body gesture generation is relatively limited.
Additionally, the high requirements for data and computation further constrain its use in CS gesture generation. 


Currently, in the field of human gesture generation, 
diffusion models \cite{ho2020denoising} are predominantly used in two main applications: generating comparably large body movements (\eg, human walking) \cite{tevet2022human,zhang2022motiondiffuse,zhao2023diffugesture,ao2023gesturediffuclip} and generating poses in Co-speech scenarios \cite{ji2023c2g2,zhi2023livelyspeaker,yang2023diffusestylegesture}.
However, the existing approaches lack the capability to tackle fine-grained gesture generation. Additionally, they primarily focus on body poses without lip movements. Lastly, their diffusion models require extensive training data, which is not feasible given the limited dataset in our CS scenarios.


\section{Method}

In this section, we provide a comprehensive description of our proposed method, GlossDiff, designed for rhythm-aware CS gesture generation, which seamlessly integrates domain-specific knowledge for CS generation. As shown in Figure \ref{fig:cuers}, our GlossDiff framework consists of three primary components: the knowledge infusion module, the rhythmic module, and the Diffusion-based generation module.

\subsection{Problem Formulation}

Automatic CS gesture generation involves generating the corresponding {\color{black} landmarks} sequence of CS gesture $M^*$, given an audio signal $A$ and the text $T$. 
In the task of automatic multi-modal CS gesture generation, the combined features of $A$, $T$, and the generated rhythmic information are input into the CS gesture generator. The final CS gestures ($M^*$) including lips, fingers, and hand positions are obtained by minimizing:
\begin{equation}
\sum_{i=1}^{L}||M_i^*-M_i||,   
\end{equation}
where $L$ represents the frame count of the current CS video. 
The ground truth CS gesture landmarks $M_{i}$ in the $i$\text{-}th frame of the CS video is obtained by the \textit{Expose} method \cite{Choutas2020expose}. 
$M_i^*=\hat{M}_i+\widetilde{M}_i$, where $\hat{M}={\rm{G_D}}(T, A)$ are generated semantic gesture landmarks representing the corresponding generated gesture in the $i$\text{-}th frame. $\rm{G_D}$ is the diffusion-based semantic gesture generator. Additionally, $\widetilde{M}={\rm{G_R}}(A)$ is the rhythmic information derived from the correspondinng audio speech, with $\rm{G_R}$ as the rhythm generator.

\subsection{Knowledge Infusion Module}
\label{subsec: Knowledge Infusion Module}
The primary objective of the knowledge infusion 
module is to transform spoken language text $T$ (\ie, the speech transcription) into direct text instructions (\ie, gloss, see Figure \ref{fig:gloss}(b)), which describe the corresponding fine-grained CS motions. To achieve this, we leverage the LLM, \ie, ChatGPT4 \cite{openai2023gpt4}, the prompt engineering approach to infuse the encoding rules of Chinese CS \cite{pilot} into our framework by the following:
\begin{equation}
g = \text{LLM}(T,P),
\end{equation}
where $P$ is our designed prompt based on CS domain knowledge (\ie, prior transformation rules of CS based on \cite{pilot}), and $T$ is the input text. Ultimately, this process enables the transformation of our indirectly semantic-related text into directly semantic-related gloss.

\subsection{Diffusion-based Generation Module}

\subsubsection{Gloss-based Motion CLIP Fine-tuning}

MotionCLIP \cite{tevet2022motionclip} is a multimodal large-scale model specifically designed for generating general motion gestures. To obtain an accurate feature embedding of CS gloss, we leverage the MotionCLIP as our pre-trained model, and fine-tune it using the generated CS gloss (introduced in Subsection \ref{subsec: Knowledge Infusion Module}) and the paired CS gestures. 


As for the fine-tuning stage, we adopt CLIP-style contrastive learning \cite{radford2021learning} to fine-tune the encoders with CS data. Given a batch of pairs containing CS gesture motion and gloss embeddings, denoted as  $\mathcal{B} =\left\{\left(z_i^m, z_i^g\right)\right\}_{i=1}^B$, where $B$ is the batch size. $\mathcal{E}_m$ and $\mathcal{E}_{g}$ are the corresponding MotionCLIP encoder for both motion sequence and gloss. $Z^m=\mathcal{E}_m(M), \quad Z^g=\mathcal{E}_g(g)$. The goal of the training is to maximize the similarity between paired $z_i^m$ and $z_i^g$ of in the batch while minimizing the similarity of the incorrect pairs $\left(z_i^m, z_j^g\right)_{i \neq j}$. A symmetric cross entropy (CE) loss $L_{CE}$ is optimized over these similarity scores. Formally, the loss is:
\begin{equation}
\begin{aligned}
\mathcal{L}_{\text {CLIP }}=\mathbb{E}_{\mathcal{B} \sim \mathcal{D}} & {\left[L_{CE}\left(\boldsymbol{y}\left(z_i^m\right), \boldsymbol{p}_{m}\left(z_i^m\right)\right)\right.} \\
& \left.+ L_{CE}\left(\boldsymbol{y}\left(\boldsymbol{z}_j^g\right), \boldsymbol{p}_{g}\left(z_j^g\right)\right)\right],  
\end{aligned}
\end{equation}
where $\boldsymbol{y}$ specifies the true correspondence between the gestures $z_i^m$ and gloss $z_j^g$ in the training batch $\mathcal{B}$. If they are paired, $\boldsymbol{y} = 1$, otherwise, $\boldsymbol{y} = 0$. $\boldsymbol{p}$ is defined as: 
\begin{equation}
\boldsymbol{p}_{m}\left(z_i^m\right)=\frac{\exp \left(z_i^m \cdot z_i^g / \eta\right)}{\sum_{j=1}^B \exp \left(z_i^m \cdot z_j^g / \eta\right)},
\end{equation}
where $\eta$ is the temperature of softmax, and $\boldsymbol{p}_{g}\left(z_j^g\right)$ follow the same computations.

\subsubsection{Gloss-Prompted Diffusion Model}
To generate CS gesture video, we propose a Gloss-Prompted Diffusion Model. More precisely, the semantic hand gesture generator $\rm{G_D}$ is designed based on the latent diffusion model \cite{Rombach2022Latent}, which applys diffusion and denoising steps in a pre-trained latent space. The latent diffusion model is trained with the standard noise estimation loss \cite{ho2020denoising} defined as: 
\begin{equation}
\label{equation5}
\mathcal{L}_{\text {noise }}=|| {\epsilon} -\epsilon_{\theta}\left(Z_n, n, g, A\right)||_2^2,
\end{equation}
where ${Z}_n$ is the latent CS gesture at each time step $n$. $A$ is audio speech, and $g$ is the generated gloss. $\epsilon$ is the ground truth noise and $\epsilon_{\theta}$ is the noise predicted by latent diffusion model, where $\theta$ is the parameters of latent diffusion model.

To inject the information of the gloss prompts into the diffusion network, we employ an adaptive instance normalization (AdaIN) layer \cite{huang2017arbitrary}. Specifically, we leverage the fine-tuned MotionCLIP gloss encoder $\mathcal{E}_g$ to convert the gloss prompt into a gloss embedding $z^g$. Then, we learn a MLP network to map the gloss embedding $z^g$ to parameters that modify the per-channel mean and variance of the AdaIn layer.

To train our Gloss-Prompted Diffusion Model, we employ classifier-free guidance as detailed in \cite{ho2022classifier}. Specifically, during training, we enable the diffusion model $\boldsymbol{G}_D$ to master both the semantic conditional and unconditional distributions by randomly configuring $\boldsymbol{g}=\varnothing$. This action effectively deactivates the AdaIN layer with a probability of $p$ during the training phase, which is set to 10\% \cite{tevet2022human}. During inference, the anticipated noise is calculated using:
\begin{equation}
\epsilon_n^*=p \epsilon_\theta\left(Z_n, n, g, A\right)+(1-p) \epsilon_\theta\left(Z_n, n, \varnothing, A\right).
\end{equation}

After obtaining the predicted noise $\epsilon_n^*$, the model operates in a reverse step-wise manner over $N$ time steps, updating a latent gesture sequence ${Z}_n$ at each time step $n$. It begins by generating a sequence of latent codes $Z_{N} \sim \mathcal{N}(0, I)$ and subsequently calculates a series of denoised sequences ${Z_{n}}$ through the iterative removal of the estimated noise $\epsilon_{n}^*$ from $Z_{n}$ ($n=N-1, \ldots, 0$). $Z_{0}$ is the final generated CS gesture latent embedding through $N$ reverse diffusion steps. $Z_{0}$ is fed into a Transformer-based decoder \cite{petrovich2021action} to generate semantic CS gesture motion $\hat{M}$. 

\subsection{Audio-driven Rhythmic Module}
In CS gesture generation, it's not just the accurate positioning of the gesture that matters; the natural rhythm of gesture motion plays a crucial role. We believe that the audio speech signal contains not only the semantic information but also the rhythmic dynamics of CS, which significantly contributes to achieving visual and auditory coherence. 

To address this, we introduce a novel Audio-driven Rhythmic Module (ARM), designed to capture the rhythmic dynamics of gestures. This module employs three convolution layers as a rhythmic dynamics generator $\rm{G_R}$, further aligning the motion dynamics with the CS rhythm.

Existing research (\eg, WavLm and AudioLDM) \cite{lebourdais2022overlapped,liu2023audioldm} have shown that compared with MFCC features, audio features extracted by the large pre-trained model have a stronger expressive capability and can avoid information loss. Without loss of generality, in this work, we use the encoder of WavLM, denoted as $\mathcal{E}_A$ to extract audio features to prevent information loss, thereby preserving richer and higher-dimensional rhythmic information.

To handle the lip-hand synchronization issue \cite{re3} in CS, we reformulate the task as one of determining the motion magnitude for each frame within consecutive motion sequences. Unlike methods that attempt to enforce perfect alignment between generated gestures and speech, our approach implicitly learns how to produce asynchronous gestures that correspond to the input speech. Rather than directly controlling the gestures of each individual frame, we focus on regulating the overall rhythm of a motion sequence.

The loss function for the ARM is defined as:
\begin{equation}
\label{equation:rhythm}
\mathcal{L}_{\text {rhythm }}
=||\widetilde{M}-\left(M-\bar{M}\right)||,  
\end{equation}
where $\bar{M}$ represents the average motion within the set of generated motions $M$. The difference between $M$ and $\bar{M}$ quantifies the magnitude of hand and finger movement. The purpose of $\mathcal{L}_{\text {rhythm}}$ is to ensure that the generated $\widetilde{M} = \rm{G_R (\mathcal{E}_A (A))}$ maintains the natural offset relative to the mean gesture. $\mathcal{E}_A$ is the encoder of WavLM. This offset helps in generating motion dynamics for a natural, non-mechanical movement without disrupting the semantics of the CS gesture. We demonstrate the efficacy regarding rhythm quality and naturalness with quantitative result in Sec. \ref{sec:quantitative}, as well as qualitative result of in Sec. \ref{sec:qualitative}.

\subsubsection{Novel Quantitative Rhythmic Metrics}
\label{sec:novel_metric}
In this work, for the first time, rhythm is investigated as an important paralinguistic feature to improve CS' communication efficacy. 
To capture the unique asynchronous dynamics between lip and hand movements in CS scenarios, we propose a novel metric, Gesture Audio Difference (GAD), to evaluate the rhythmic synchronization of the generated gestures.
This metric is defined as follows:
\begin{equation}
\operatorname{GAD}(M, A)=\frac{1}{N} \sum_{i=1}^{N} \textbf{1}[||\boldsymbol{U}_i^M-\boldsymbol{U}_i^A||_1<\tau],
\end{equation}
where $M$ and $A$ represent the CS gesture and audio speech, respectively. The term $N$ denotes the number of annotated temporal segments, which are equal for both speech and gesture. The variable $\boldsymbol{U}_i$ refers to the middle time instant of a segment, indicating a specific moment when a gesture or speech occurs. The function \textbf{1} is an indicator function, mapping elements within the subset (satisfying $||\boldsymbol{U}_i^M-\boldsymbol{U}_i^A||_1<\tau$) to one, and all other elements to zero.

Taking the asynchrony between audio speech and CS hand movements into consideration, we introduce a threshold $\tau$, which ensures their alignment and is empirically determined based on a statistical study of the hand preceding time \cite{liu2020re}.






\subsection{Training of GlossDiff Framework}
We employ a semantic loss to ascertain the semantic accuracy of the final generated gestures. To be specific,
\begin{equation}
\label{equation:semantic}
\mathcal{L}_{\text {semantic }}=1-\cos \left(Z_0, Z_0^{ *}\right),
\end{equation}
where $\cos (\cdot, \cdot)$ represents the cosine distance, while $Z_0$ and $Z_0^{*}$ denote the final generated CS gesture latent embedding and the ground truth CS gesture motions, respectively. 



Following the existing training procedure for denoising diffusion models, we optimize the following loss:
\begin{equation}
\mathcal{L}_{\text {total }}= \alpha \mathcal{L}_{\text {noise }}+ \beta \mathcal{L}_{\text {semantic }}+\gamma \mathcal{L}_{\text {rhythm }},
\end{equation}
where $\alpha$ is the weight of $\mathcal{L}_{\text {noise }}$ (in Equation (\ref{equation5})), $\beta$ is the weight of $\mathcal{L}_{\text {semantic }}$
(in Equation (\ref{equation:semantic})), and $\gamma$ is the weight of $\mathcal{L}_{\text {rhythm }}$ (in Equation (\ref{equation:rhythm})). 

\begin{table*}[!t]
   \begin{center}
  \caption{Experiment results on MCCS Dataset compared with SOTA methods. ``Gloss-Prompt" indicates the integration of a Gloss Knowledge Infusion Module. The term ``WavLM" refers to the substitution of MFCC features with features from the pre-trained large-scale speech model, wavLM. ``Gloss-CLIP" denotes the incorporation of Gloss-based Motion CLIP Fine-tuning. }
  \label{tab:mccshi}
  \setlength{\tabcolsep}{3mm}{
  \begin{tabular}{lccccc}
    \toprule
    Methods & {PCK (\%)$\uparrow$}& {FGD$\downarrow$} &{MAJE ($mm$)$\downarrow$} & {MAD ($mm/s^2$)$\downarrow$ }&{GAD (\%)$\uparrow$}\\
      \hline
      Speech2Gesture \cite{ginosar2019learning}   &36.84& 19.25 & 61.26 & 3.97&66.8\\
      GTC \cite{GTC}        &41.23& 6.73&55.43&2.54&66.7\\
      HA2G \cite{liu2022learning}           &43.51&4.07& 46.78&2.29& 67.2\\
      DiffGesture \cite{zhu2023taming}     &47.58&\textbf{3.50}&48.52&2.12&69.9\\
      \midrule
      Our GlossDiff (w/o Gloss-prompt)              &51.12&4.72&  45.68 & 1.28&75.6  \\
      Our GlossDiff (w/o WavLM)              &52.97&4.54& 42.31 & 0.71&78.3  \\
      Our GlossDiff (w/o Gloss-CLIP)            &53.41&4.31& 43.52 & 0.65&79.1\\
      Our GlossDiff                      &\textbf{54.23}&3.92& \textbf{39.28} & \textbf{0.52}&\textbf{79.4}  \\
      \hline
  \end{tabular}}
    \end{center}
\end{table*}



\section{Experiments}



\subsection{MCCS Datatset}
\label{MCCS-HI}
Previously, only two CS datasets were available for public access: One was in French\footnote{\url{https://zenodo.org/record/5554849\#.ZBBCvOxBx8Y}} \cite{liu2018visual}, consisting of recordings of a single cuer delivering 238 sentences; The other was in British English\footnote{\url{https://zenodo.org/record/3464212\#.ZBBAJuxBx8Y}} \cite{liu2019automatic}, similarly featuring a single cuer reciting 97 sentences. 
To remedy the scarcity of Chinese CS data, 
we built in this work, for the first time, a large-scale \textbf{M}andarin \textbf{C}hinese \textbf{CS} dataset that includes contributions from four CS cuers, called \textbf{MCCS}.

We first select 1000 text sentences following the below principles: (1) They cover common scenarios in daily life, including colloquial dialogues, more formal words, as well as written words. (2) The materials aim to cover possible syllable combinations.
All in all, our text album covers 23 main topics, 72 subtopics, and the most commonly used 399 Mandarin syllables. It comprises a total of 1000 sentences, 10,482 words with an average of 10.5 words per sentence. The shortest sentence contains 4 words, while the longest has 25 words. Then, we recorded CS videos for each of the four cuers performing the 1000 sentences, resulting 4000 sentences in total.


All videos are recorded using either a camera or a mobile phone in landscape mode, 
The four cuers have received systematic training to ensure they can perform Mandarin Chinese CS smoothly and accurately. 
Note that our dataset has been collected with the explicit consent of the individuals involved and is eligible for open source.

\subsection{Experimental Setup}
During the training phase, we pre-train the motion clip first and then follow an end-to-end pipeline to train the latent diffusion model. The experiments are implemented using PyTorch, with four \textit{A6000} GPU cards for model training. During the inference phase, we use the latent diffusion model to generate CS gestures. The training and test data are randomly split as $4:1$. The number of diffusion steps is 1000, and the training batch size is 128. The weight of loss items is set to $\alpha=1$, $\beta=0.2$ and $\gamma$ = 0.1. 


\subsubsection{Evaluation Metrics}
The conventional evaluation metrics of the generated gestures contain three classes: Percentage of Correct Keypoint (PCK) \cite{PCK}, Fréchet Gesture Distance (FGD) \cite{GTC}, Mean Absolute Joint Errors (MAJE) \cite{GTC}, and Mean Acceleration Difference (MAD) \cite{GTC}. In addition, to further measure the unique asynchronous dynamics between lip and hand movements in CS scenarios, we use the novel metric GAD as described in Sec.\ref{sec:novel_metric} to evaluate the rhythmic synchronization of the generated gestures.

\subsection{Quantitative Result and Analysis}
\label{sec:quantitative}
\subsubsection{Comparison with SOTA}
We compare our approach with four recent gesture synthesis methods, \ie, Speech2Gesture \cite{ginosar2019learning}, Gestures from Trimodal Context (GTC) \cite{GTC}, HA2G \cite{S2AG}, DiffGesture \cite{zhu2023taming}. 
We take DiffGesture as the SOTA method among these approaches, since it achieves the best result on the TED Gesture datasets \cite{yoon2019}. 

Table \ref{tab:mccshi} provides a detailed comparison among our methods and the previous methods on the MCCS datasets. Our method GlossDiff gives the best results in PCK, MAJE, MAD, and GAD metrics, most of which have a wide superiority leap comparing to the reference systems. The results demonstrate a higher quality of fine-grained gesture generation by our proposed system. The only exception is one FGD score that slightly trails the SOTA method, while it surpasses all other reference methods. Notably, our method's PCK values are significantly higher than other methods, showing its effectiveness in fine-grained generation. Moreover, our method excels in rhythm performance, achieving the highest GAD values. This superiority on GAD metrics demonstrates that our method can effectively capture the rhythm in CS gesture.

\subsubsection{Ablation Study}
We provide the ablation study for three modules in Table \ref{tab:mccshi}. The term ``Gloss-prompt" indicates the integration of a Gloss Knowledge Infusion Module. ``WavLM" refers to using features extracted from the pre-trained large-scale speech model wavLM instead of conventional MFCC. ``Gloss-CLIP" denotes the incorporation of Gloss-based Motion CLIP Fine-tuning. We can observe that the absence of any module leads to a decline in performance metrics, demonstrating the efficacy of each module in our framework. Specially, the absence of the Gloss-prompt and Gloss-CLIP modules results in a decrease in PCK by 1.85\% and 1.26\%, respectively, highlighting their critical role in fine-grained generation. 

\begin{figure}[htb]
\begin{minipage}[b]{1.0\linewidth}
 \centering
 \centerline{\includegraphics[width=8.5cm]{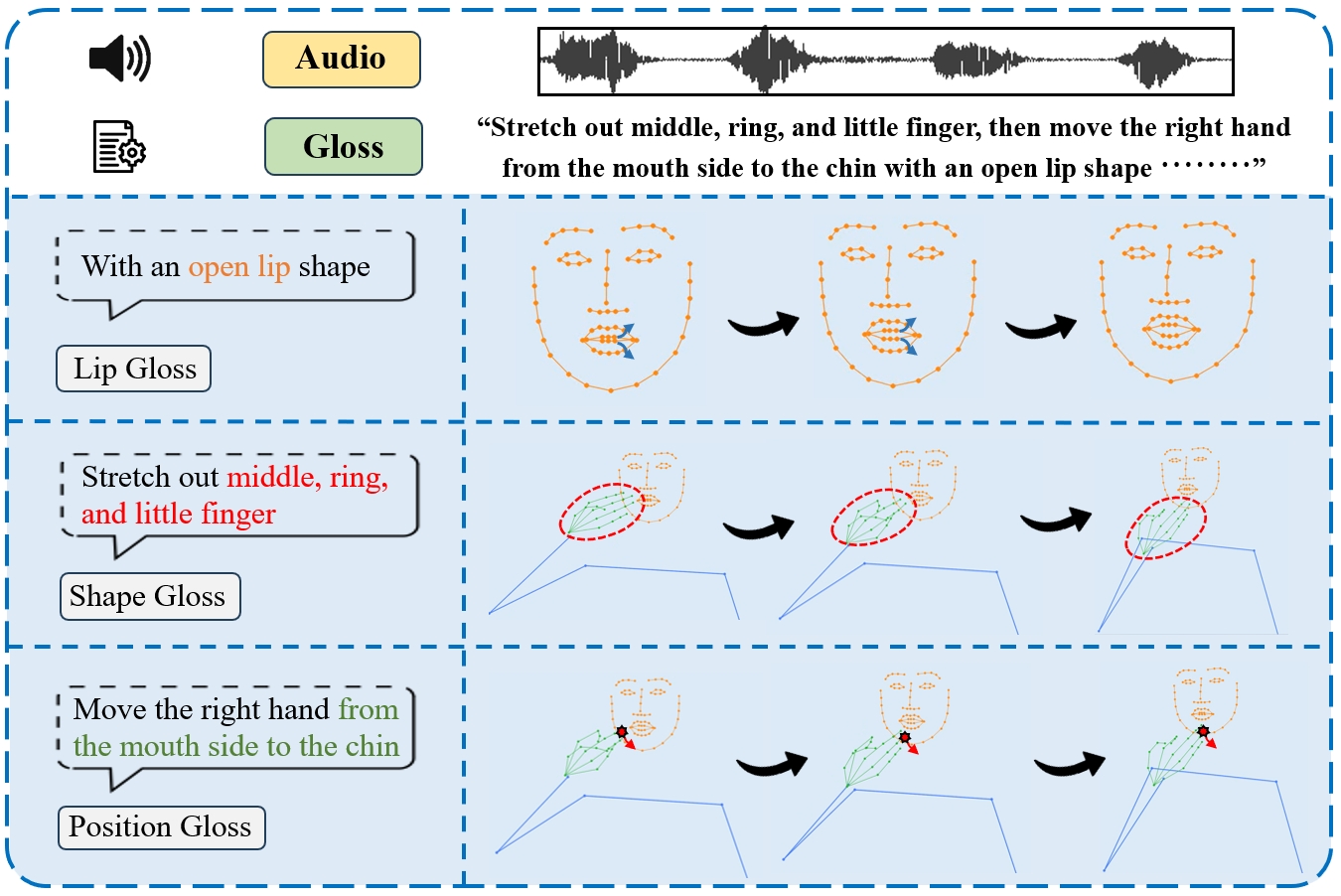}}
\end{minipage}
\caption{The visualization result of the generated gesture according to fine-grained Gloss. Better view by zooming in.}
\label{fig:vis-gloss}
\end{figure} 

\subsection{Qualitative Result and Analysis}
\label{sec:qualitative}
\subsubsection{Visualization of Generated Fine-grained CS Gesture}
Figure \ref{fig:vis-gloss} shows fine-grained hand gestures generated with gloss prompts, where each row shows the detailed gloss of different body parts and their gesture sequences. We used arrows to indicate lip movement trends, red circles for finger shape transformations, and red stars for hand position shifts, including their movement directions. The first row in the figure shows the lips' contour expanding as the gloss input. The second row emphasizes finger shape changing aligned with detailed finger gloss. In the third row, there are subtle hand position shifts, marked by red stars moving from near the mouth to the chin area, showing our method's effectiveness in using detailed gloss to guide CS gesture generation.

\subsubsection{Distribution of Fine-grained Gesture Feature}
\begin{figure}[htb]
\begin{minipage}[b]{1.0\linewidth}
 \centering
 \centerline{\includegraphics[width=8.5cm]{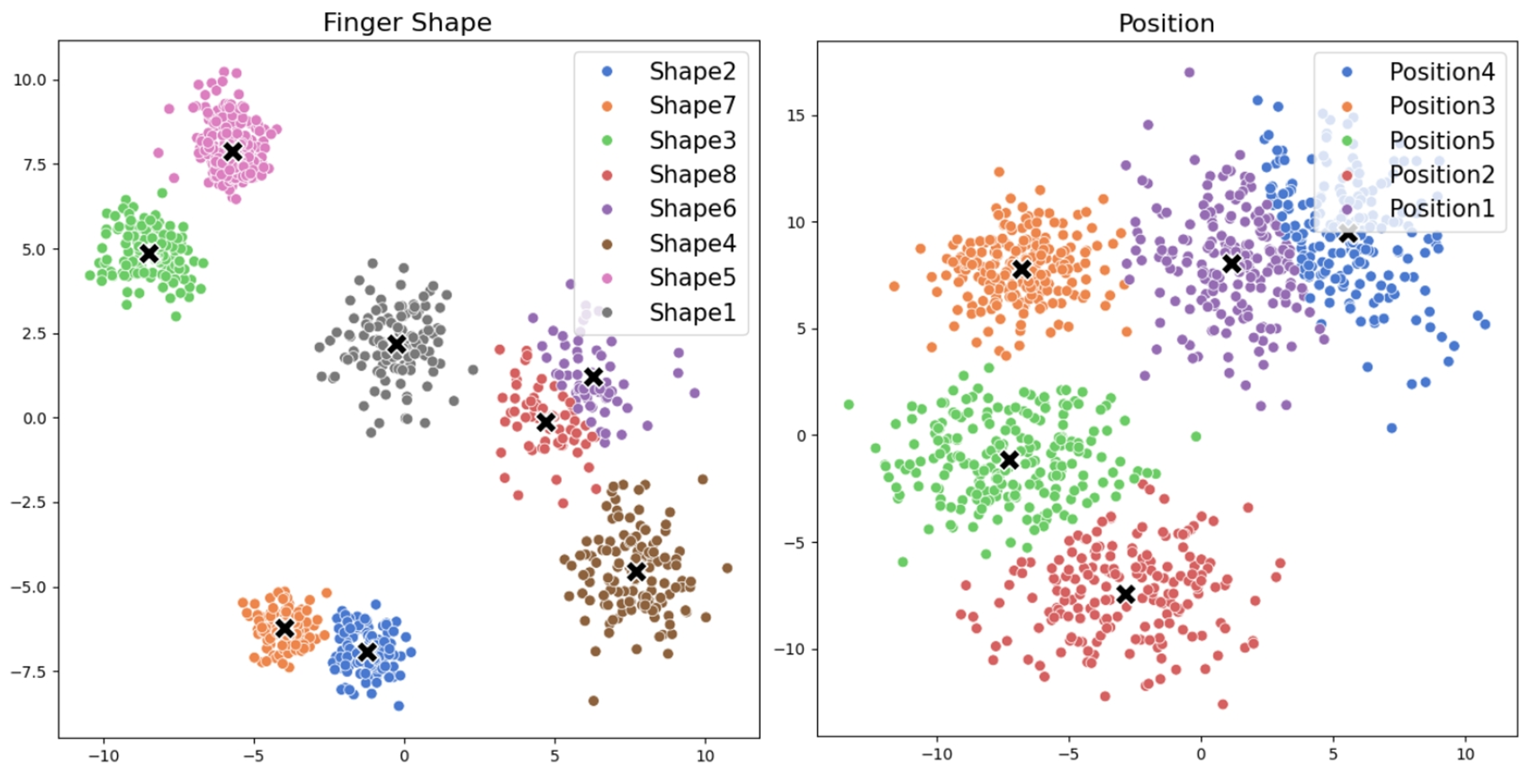}}
\end{minipage}
\caption{The visualization of t-SNE clustering for eight groups of consonants
corresponding to finger shapes, and five groups of vowels corresponding to hand position. Each color represents a
group of consonants or vowels.}
\label{fig:vis-tsne}
\end{figure}

To visualize the generated CS gesture in the feature space, we used t-SNE \cite{tsne} for dimension reduction. We uniformly select frames from the generated CS sequences and extract the hand gesture features corresponding to the text. Recall that, as depicted in Figure \ref{fig:gloss}, the MCCS incorporates 8 distinct finger shapes to signify the 24 consonants of the Chinese language, along with 5 hand positions to denote the 16 vowels. In the left part of Figure \ref{fig:vis-tsne}, the 8 distinct clusters are separate, with each cluster corresponding to a set of finger shapes (where each color represents a different consonant group). Some clusters that are very close in distance have similar finger shapes, such as shape8 and shape6, as well as shape2 and shape7. This visualization validates the effectiveness of our method in capturing the fine-grained semantics of CS hand and finger shapes. On the right side of Figure \ref{fig:vis-tsne}, We can find different hand positions have differences in features, but there is more overlap among the clusters,  which means they are not as distinctly differentiated in feature-level as finger shapes. 

\subsubsection{Visualization of Generated CS Gestures}
\begin{figure}[htb]
\begin{minipage}[b]{1.0\linewidth}
 \centering
 \centerline{\includegraphics[width=9.0cm]{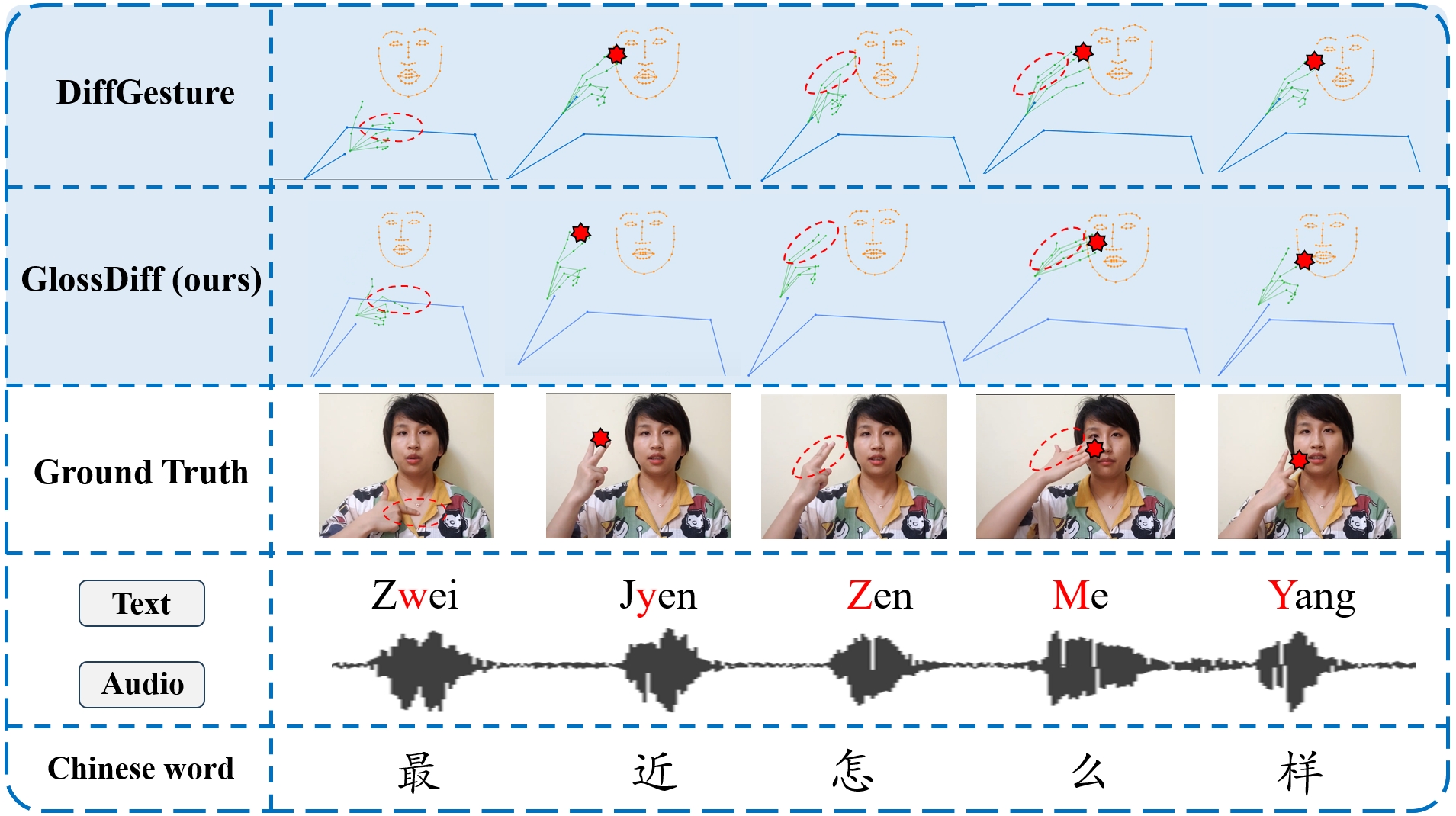}}
\end{minipage}
\caption{The visualization result of the generated gestures compared to SOTA method. Better view by zooming in.}
\label{fig:vis-sota}
\end{figure}

Figure \ref{fig:vis-sota} compares the visualization results of our method with the SOTA method, DiffGesture. This comparison includes the gestures' corresponding audio, text, and ground truth video frames. We highlighted corresponding phonemes in red and used red stars and circles to indicate hand locations and finger shapes, respectively.

Our method shows a noticeable improvement in gesture accuracy, particularly in fine-grained details. For example, our index finger shape is more precise than the SOTA method, as seen in the first column. In the second column, our method accurately places the hand beside the face, unlike the SOTA method's placement beside the eye. The fourth column illustrates our method's superior precision in thumb position and overall gesture alignment with the ground truth, showing greater adherence to CS rules and enhanced detail accuracy.

\begin{figure}[htb]
\begin{minipage}[b]{1.0\linewidth}
 \centering
 \centerline{\includegraphics[width=8.5cm]{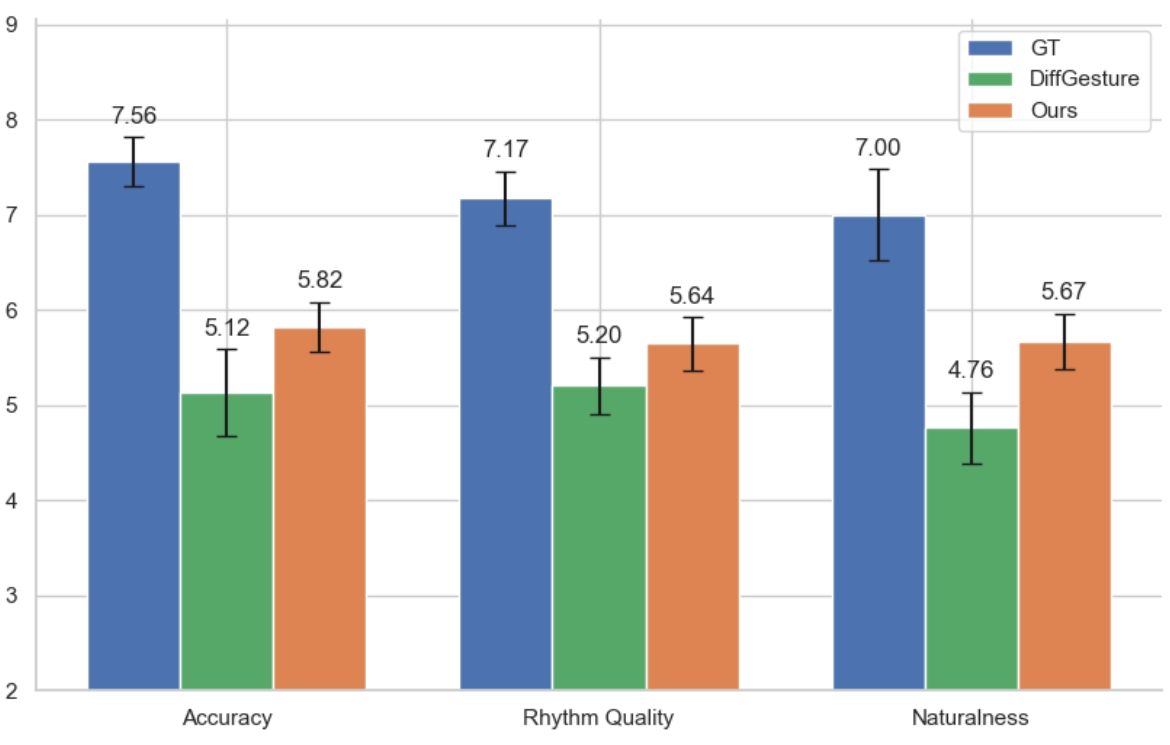}}
\end{minipage}
\caption{User study results of the ground truth (GT)), current SOTA (DiffGesture) and our method (GlossDiff).}
\label{fig:user}
\end{figure}

\subsubsection{User Study}
We conduct a user study to evaluate CS gestures generated by our method compared with SOTA and the ground truth.  
This study involved 10 groups of videos, each with a ground-truth CS gesture video, videos generated by the current SOTA method (DiffGesture) and our method (GlossDiff). All videos were randomly shuffled. Ten subjects trained in CS were asked to rate the CS gesture videos from three perspectives: accuracy, rhythm quality, and naturalness, each with a score ranging from 0 to 10 (the higher the better). We calculated average scores and confidence intervals for each case.

It is shown in Figure \ref{fig:user} that our method surpassed the current SOTA DiffGesture in all three metrics, getting closer to the ground truth. This demonstrates our method's ability to produce more accurate and natural CS gestures, especially in rhythm quality, attributed to the proposed ARM. Our approach notably outperforms the DiffGesture in accuracy, proving its effectiveness in fine-grained gesture generation.

\section{Conclusion}
We introduced a novel GlossDiff framework that effectively generates fine-grained CS gesture sequences. We have proposed a gloss knowledge infusion module and an audio rhythm module for an accurate and natural CS gesture video generation. Additionally, we contributed the first large-scale MCCS dataset. Extensive experiments on MCCS demonstrate our approach's efficacy, surpassing current SOTA methods. Qualitative experiments and ablation studies validated our system's overall effectiveness as well as each individual module's. Future work aims to infuse CS video generation with prosody and emotion. The Automatic Prompt Engineering (APE) is also a promising direction to improve gloss quality.

\section{Acknowledgement}
This work was supported by the National Natural Science Foundation of China (No. 62101351), and  Guangzhou Municipal Science and Technology Project: Basic and Applied Basic research projects (No. 2024A04J4232).

\bibliographystyle{named}
\bibliography{ijcai24}

\end{document}